# Visualization and Analysis of Frames in Collections of Messages:

# Content Analysis and the Measurement of Meaning

**Esther Vlieger and Loet Leydesdorff** [a]

University of Amsterdam, Amsterdam School of Communication Research (ASCoR)
Kloveniersburgwal 48, 1012 CX Amsterdam, The Netherlands.

**Abstract**

A step-to-step introduction is provided on how to generate a semantic map from a collection of messages (full texts, paragraphs or statements) using freely available software and/or SPSS for the relevant statistics and the visualization. The techniques are discussed in the various theoretical contexts of (i) linguistics (e.g., Latent Semantic Analysis), (ii) sociocybernetics and social systems theory (e.g., the communication of meaning), and (iii) communication studies (e.g., framing and agenda-setting). We distinguish between the communication of information in the network space (social network analysis) and the communication of meaning in the vector space. The vector space can be considered a generated as an architecture by the network of relations in the network space; words are then not only related, but also positioned. These positions are *expected* rather than observed and therefore one can communicate meaning. Knowledge can be generated when these meanings can recursively be communicated and therefore also further codified.

**Keywords**: semantic map, measurement of meaning, frame, discourse, communication studies

[a] Corresponding author: loet@leydesdorff.net ; http://www.leydesdorff.net.

## 1. Introduction

The study of latent dimensions in a corpus of electronic messages has been part of the research agenda from different disciplinary perspectives. In linguistics, for example, these efforts have been focused under the label of "latent semantic analysis" (LSA; Landauer *et al.*, 1998); in communication studies, "framing" is a leading theoretical concept for studying the latent meanings of observable messages in their contexts (e.g., Scheuffele, 1999); and in social-systems theory and socio-cybernetics, codes of communication which can be symbolically generalized (Parsons, 1963a and b; 1968; Luhmann, 1995 and 2002; Leydesdorff, 2007) are expected to operate latently or virtually as "a duality of structure" (Giddens, 1984; Leydesdorff, 2010). These efforts have in common that the analyst shifts his/her attention from the communication of information in observable networks to the communication of meaning in latent dimensions.

Latent dimensions can be operationalized as the "eigenvectors" of a matrix representing the network under study. Eigenvectors, however, operate in a vector space that can be considered as the architecture spanned by the variables (vectors) in observable networks. Statistical techniques for analyzing latent dimensions such as factor analysis and multi-dimensional scaling (MDS) are well-known to the social scientist—and where further developed for the purpose of analyzing communication (Lazarsfeld & Henry, 1968)—but the current enthusiasm for network analysis and graph theory has tended to push aside these older techniques in favour of a focus on observable networks and their structures. Spring-embedded algorithms that operate on networks such as Kamada & Kawai (1989) or Fruchterman & Reingold (1991) are integrated in software packages freely available at the internet such as Pajek and Gephi. These newer visualization capacities far outreach the traditional ones such as MDS.[1]

In this introduction, we show how one can use these newer visualization techniques with the older factor-analytic approach for distinguishing main dimensions in order to visualize the communication of meaning as different from the communication of information. The communication of information can be considered as the domain of social network analysis and its semantic pendant in traditional co-word analysis (Callon *et al.*, 1983; 1986). Words and co-words, however, cannot map the development of the sciences (Leydesdorff, 1997). The architectures of the discourse have first to be analyzed and can then also be visualized. Using an example, we walk the user through the different steps which lead to a so-called Pajek-file which can be made input to a variety of visualization programs.

In other words, we provide a step-by-step introduction that enables the user to generate and optimize network visualizations in the vector space, that is, the space in which meaning is communicated as different from the communication of information in the network. Meaning can be generated when informations are related at a systems level. In cybernetics, one often invokes an "observer" to this end (Edelman, 1989; Von Foerster, 1982), but a discourse can also be considered as a relevant system of reference. Note that meaning is provided in terms of expectations and can be informed and updated by observations. The various bits of informations can be positioned in a vector space in addition to being related or not in terms of network relations (Burt, 1982). The absence of relations can then be as informative as their presence (Burt, 1995; Granovetter, 1973).

[1] VosViewer, a visualization program available at http://www.vosviewer.org, reads Pajek files as input, but uses an algorithm akin to MDS (Van Eck *et al.*, 2010).

The software for the visualization and animation of the vector space uses the cosine values of the angles between the vectors (variables) of word occurrences in distributions. We explain below how the word-document matrix can additionally be used as input to the factor analysis; for example, in SPSS. Unlike "single value decomposition" (SVD) which has been the preferred method in latent semantic analysis, factor analysis is available in most social-science statistics programs. We developed software so that one can move from a set of textual messages (either short messages or full texts) to these different software packages and take it further from there.

**2. The framing concept**

The concept of framing was introduced by Goffman (1974). He explained that messages in the mass media are "framed," which means that a description is provided from a certain perspective and with a specific interpretation. McCombs (1997) described framing as "the selection of a restricted number of thematically related attributes for inclusion on the media agenda when a particular object is discussed" (pp. 297-298). Van Gorp (2007) indicated that this process of selection is inevitable, as journalists are unable to provide an objective image of reality. McQuail (2005, at p. 379) agreed on this inevitability, which results in the inclusion of a specific way of thinking into the process of communication. Entman (1993) also argues that this process of selection can either be conscious or unconscious. A certain way of thinking is transmitted through the text. As Entman (1993, at p. 52) argued:

> Framing essentially involves selection and salience. To frame is to select some aspects of a perceived reality and make them more salient in a communicating text, in such a way as to promote a particular problem, definition, causal interpretation, moral evaluation, and/or treatment recommendation for the item described. (p. 52)

A fact never has a meaning in itself, but it is formed by the frame in which it is used (Gamson, 1989). This latent meaning is implied by focusing on certain facts and by ignoring others. Frames appear in four different locations in the communication process: at the sender, within the text itself, with the receiver, and within culture (Entman, 1993).

When studying frames through the methods described below, one focuses on the frames that are embedded within the texts. These frames are often powerful, as changing a specific frame by a source might be interpreted by relevant audiences as inconsistent or unreliable (since dissonant). Textual frames are formed, among other things, by the use of certain key words and their relations. The relations among keywords provide the basis for this methodology.

Matthes and Kohring (2008) distinguished five methodological approaches to the measurement of media frames. First, in the qualitative *Hermeneutic approach,* frames are identified by providing an interpretative account of media texts linking up frames with broader cultural elements. Secondly in the *Linguistic approach,* one analyzes the selection, placement, and structure of specific words and sentences in a media text. A third model is provided by the *Manual holistic approach*. Frames are generated through a qualitative analysis of texts, after which they are coded in a manual quantitative content analysis. In the *Deductive approach*, fourthly, frames are theoretically derived from the literature and then coded in a standard content analysis. Lastly, the authors identify a fifth and methodological approach: the *Computer-assisted approach*. An example of this latter approach is elaborated in this study. In this approach, frame mapping can be used as a method of finding particular words that occur together. As frames then are generated by computer programs, instead of being "discovered" by the researcher(s), this method has the advantage of being a more objective tool for frame extraction than the other methods.

## 3. The dynamics of frames

Through the research methods described in this study, one is able to study differences or changes in frames within different discourses, not only statically, but also dynamically. Danowski (2007) studied changes in frames from the perspective of language. He indicated that frames relate to the way that facts are characterized, which is based on cultural and social backgrounds. This is consistent with the vision of Scheufele (1999), who stated that the influence of media on the public mainly works through transferring the importance of specific aspects about a certain issue. Framing is considered by Danowski (2007, 2009) as a way of shaping the process of agenda setting. He also states that framing is mainly applied to provide a positive or negative view on an issue. Unlike Entman (1993), Danowski (2007) argued that frames change relatively rapidly in the media. Discourse in the public domain would have a character more versatile and volatile than scholarly discourse. This contrast makes it interesting to study changing frames within specific discourses.

Our research method provides also a way of studying these possible differences in the dynamics. The existing network visualization and analysis program Visone was further developed for this purpose with a dynamic extension (at http://www.leydesdorff.net/visone). The network files for the different years can be assembled with a routine mtrx2paj.exe which is available with some instruction from http://www.leydesdorff.net/visone/lesson8.htm. An in-between file (named "pajek.net") can be harvested and also be read by other network animators such as SoNIA: Social Network Image Animator or PajekToSVGAnim.Exe (Leydesdorff *et al.*, 2008). In this study, however, we limit ourselves to the multi-variate decomposition of a semantic network in a static design (including comparative statics). An example of the potential of the dynamic extension to Visone showing heterogeneous networks in terms of their textual representations ("actants") can be found at http://www.leydesdorff.net/callon/animation/ which was made for a *Liber Amicorum* at the occasion of the 65th birthday of Michel Callon (Leydesdorff, 2010b).

When analyzing frames, one can make a distinction between restricted and elaborated discourses. Graff (2002) indicates that this distiction is mostly related to the audience of the communication. In restricted discourse, one single and specific meaning is constructed and reproduced. This is, for example, important in scholarly communication, when the audience in a particular field of studies has specific knowledge on the topics of communication. In elaborated discourse, communication is aimed at a wider audience. In this case, multiple meanings are created and translated into one another. The visualization of the frames in the collection of messages to be analyzed can reveal a more elaborated versus a more restricted type of discourse (Leydesdorff & Hellsten, 2005).

## 4. Using semantic maps for the study of frames

The research method presented in this section deals with content analysis of collections of messages. In addition to manual content analysis (Krippendorff, 1980; Krippendorff & Bock, 2009), one can use computer programs to generate semantic maps on the basis of large sets of messages. A properly normalized semantic map can be helpful in detecting latent dimensions in sets of texts. By using statistical techniques, it is possible to analyze the structure in semantic networks and to color them accordingly.

Content can be contained in a set of documents, a sample of sentences, or any other textual units of analysis. In our design, the textual units of analysis will be considered as the cases, and the words in these messages—after proper filtering of the stopwords—as the variables. Thus, we operate with matrices. Matrices which contain words as the variables in the columns and textual units of analysis as cases in the rows (following the convention of SPSS) are called word/document matrices. In co-word analysis and social network analysis, one often proceeds to the symmetrical co-occurrence matrix, but this latter matrix contains less information than the asymmetrical word/document matrix (Leydesdorff & Vaughan, 2006).

When visualizing a word/document matrix, a network appears, containing the interrelationships among the words and the textual units. In order to generate this network, one needs to go through various stages using different programs. In this section we explain how to generate, analyze, and visualize semantic maps from a collection of messages using the various programs available at http://www.leydesdorff.net/indicators and standard software such as SPSS and Pajek.

### *4.1. Generating the word/document occurrence matrix*

In order to generate the word/document occurrences matrix, one first needs to save a set of messages in such a format that the various programs to be used below are able to use them as input files. If the messages are short (less than 1000 characters), we can save them as separate lines in a single file using Notepad or another text editor.[2] This file has to be called "text.txt". In this case one can use the program Ti.exe (available at http://www.leydesdorff.net/software/ti) that analyzes title-like phrases. If the messages are longer, the messages need to be saved as separate text-files, named text1.txt, text2.txt, etc.[3] These files can be read by the program FullText.exe (at http://www.leydesdorff.net/software/fulltext/).

#### *4.1.1. Frequency List*

The text-file text.txt can directly serve as input for the program Frequency List (FrqList.exe at http://www.leydesdorff.net/software/ti). This program produces a word frequency list from the file text.txt, needed for assessing which words the analyst wishes to include in the word/document occurrences matrix. As a rule of thumb, more than 75 words are difficult to visualize on a single map, and more than 255 variables may be difficult to analyze because of systems limitations in SPSS and Excel 2003.

---

[2] If one uses Word or WordPad, one should be careful to save the file as a so-called DOS plain text file. When prompted by Word, choose the option to add CR/LF to each line. (CR/LF is an old indication of Carriage returns and Line feeds, like using a typewriter.)

[3] Sometimes, Windows adds the extension .txt automatically. One should take care not to save the files with twice the extension ".txt.txt". The programs assume only a single ".txt" and will otherwise lead to an error.

Together with the text-file, one can use a standard list of stopwords in order to remove the irrelevant words directly from the frequency list. It can be useful to check the frequency list manually, to remove potentially remaining stopwords. If we begin with long texts in different files (text1.txt, text2.txt, … etc.),[4] these files have first to be combined into a single file text.txt that can be read by FrqList, for the purpose of obtaining a cumulative word frequency list across these files.[5] The use of FrqList is otherwise strictly analogous.

To be able to run FrqList, one needs to install the program in a single folder with the Text-file with all the messages (text.txt) and the list of stopwords, as shown in Figure 1.

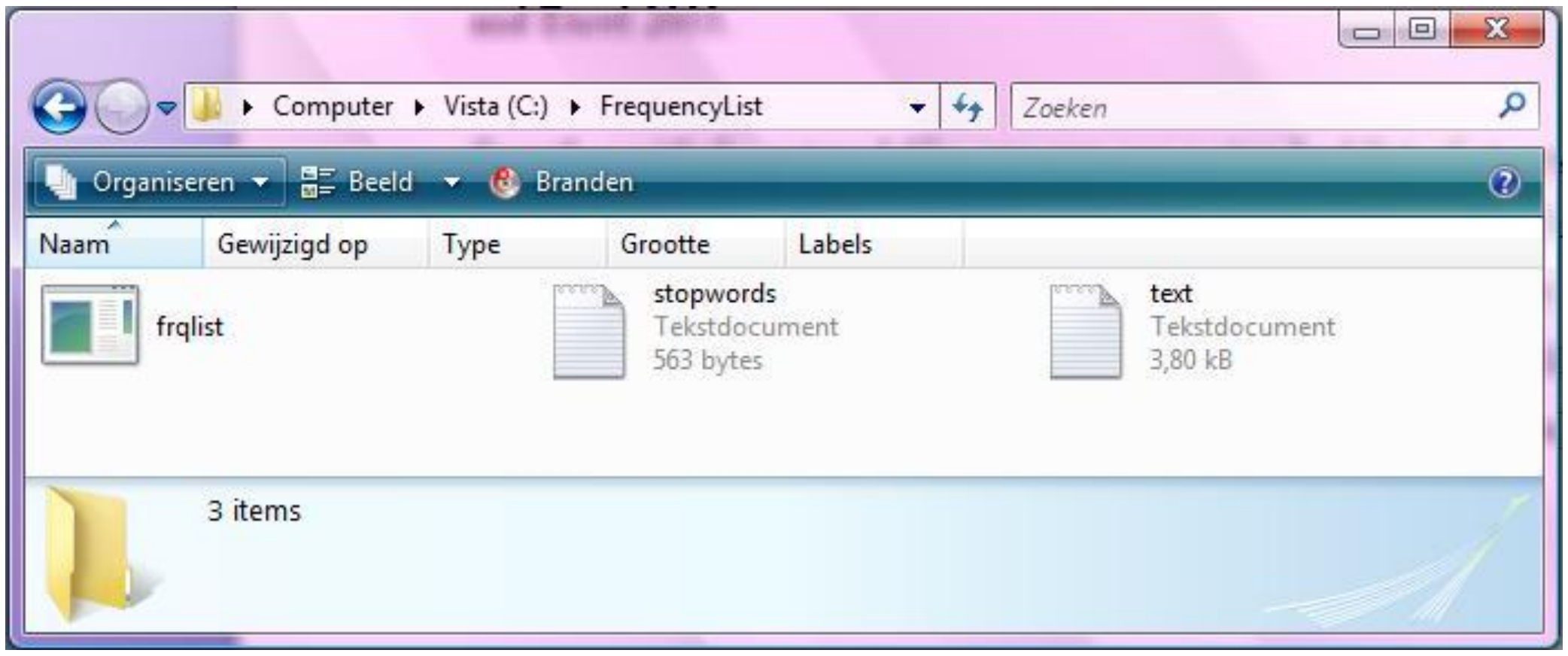

***Figure 1*** Example of using FrqList

[4] Sample files text1.txt, text2.txt, text3.txt, text4.txt can be found at http://www.leydesdorff.net/software/fulltext/text1.txt , etc.
[5] One can combine these files in Notepad or alternatively by opening a DOS box. In the DOS box, use "cd" for changing to the folder which contains the files and type: "copy text*.txt text.txt". Make sure to erase an older version of text.txt first.

After running, the program FrqList produces a frequency list: the combined word frequency list is made available as WRDFRQ.txt in the same folder, as can be seen in Figure 2. This file can be read into Excel in a next step so that, for example, the single occurrences of words can be discarded from further analysis.

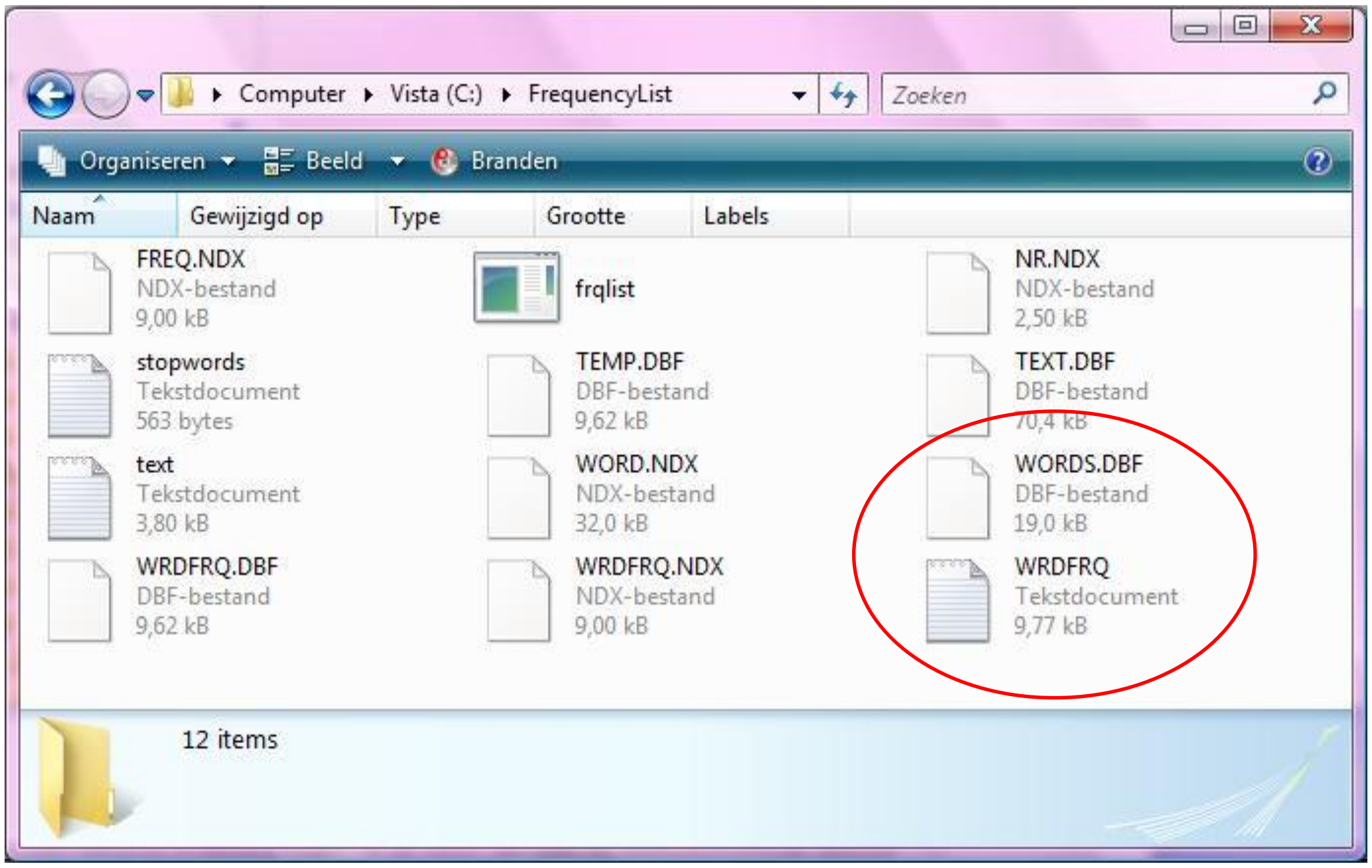

***Figure 2*** Output FrqList

*4.1.2. Full Text*

The next step is to import the frequency list—one can use wrdfrq.dbf or wrdfrq.txt—into an Excel file in order to separate the words from the frequencies as numerical values. At this stage, the list of words may be too long to use efficiently. To be able to visually interpret the data at a later stage, it can be advised to use a maximum of approximately 75 words. The first 75 words from the frequency list (without the frequencies) need to be saved as a Text-file by the name of

words.txt. (Use Notepad for saving or obey the conventions for a plain DOS text as above.) This file "words.txt" can serve as input for the programs Ti.exe or FullText.exe.

One can use ti.exe for the case that the texts are short (< 1000 characters) and organized as separate lines in a single file text.txt, but fulltext.exe is used in the case of a series of longer text files named text1.txt, text2.txt, text3.txt, …, etc. Both programs need in addition to the information in the textual units, an input file named words.txt (in the same folder) with the information about the words to be included as variables. Prepare this file carefully using the instructions about removing stopwords and making selections specified above. You may wish to run FrqList.exe a second time with a manually revised file stopword.txt. (Save this file as a DOS file!)

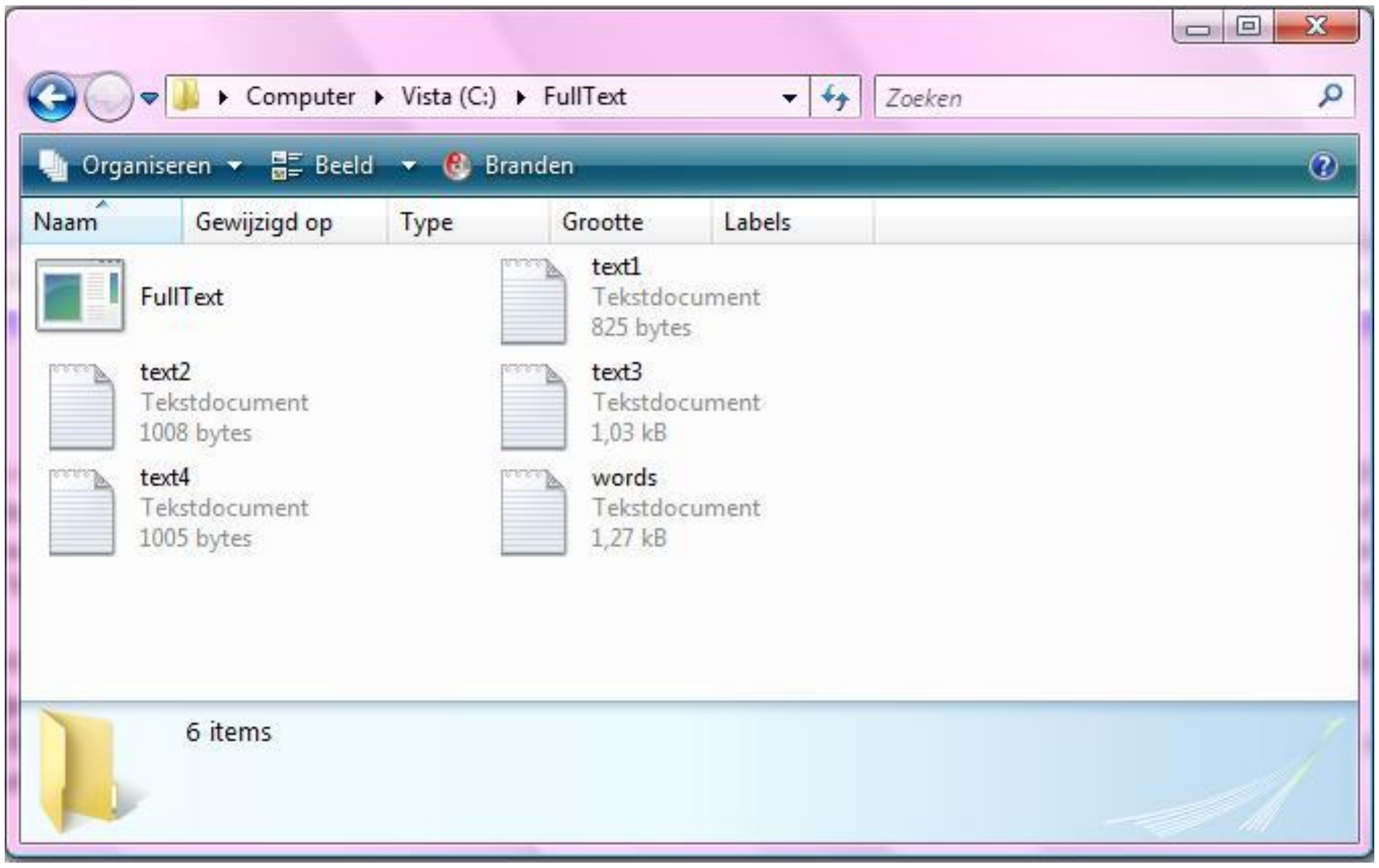

***Figure 3*** Example FullText

As can be seen in Figure 3, the separately saved messages (text1.txt, text2.txt, etc.), together with the file words.txt, form the input for FullText. (Analogously, for Ti.exe one needs the files

text.txt and words.txt.) The program produces data files, which can be used as input for the statistical program SPSS and the network visualization program Pajek. By installing the program FullText in the same folder containing the saved messages and words.text, the program can be run. The output of FullText can also be found in this same folder, as can be seen in Figure 4.

Prior to running FullText, the program demands to insert the file name ("words") and the number of texts. After running FullText (or Ti.exe), one can use the files matrix.txt[6] and labels.sps to statistically analyze the word/document occurrence matrix by using SPSS. (The file matrix.txt contains the data and can be read by SPSS. The file labels.sps is an SPSS syntax file for labelling the variables with names.) In order to generate a visualization of the semantic map, one can use the file cosine.dat as input to Pajek. How to use these files for Pajek and SPSS will be discussed in the next paragraph.

[6] Matrix.txt contains the same information as matrix.dbf. Matrix.dbf can directly be used with more than 256 variables in Excel 2007 and higher, but not in lower versions. In SPSS this depends on the version.

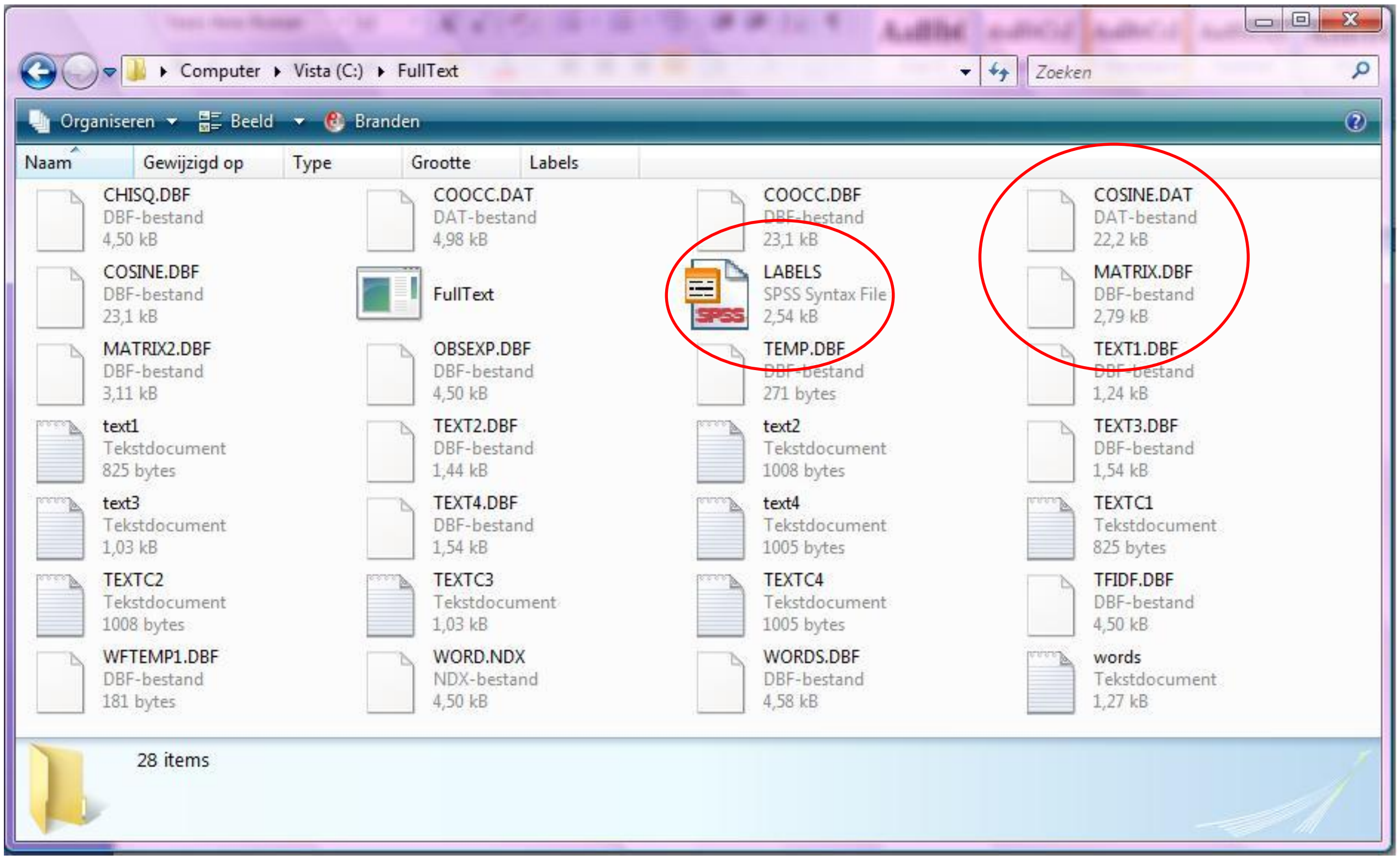

***Figure 4*** Output FullText

*4.2. Analyzing the word/document occurrence matrix*

As noted, the file matrix.txt can be read by SPSS. To label the variables with names, choose “File – Open – Syntax” in order to read the file labels.sps. Choose “Run – All”. As can be seen in the syntax file, FullText has deleted the “s” at the end of the words. The aim is to remove the plural forms, but this may have no use when analyzing a word/document occurrence matrix. By comparing to the original words in the file WRDFRQ.txt (which was generated by FrqList) the labels in the variable view of SPSS can be manually adapted. This is only necessary if one wants to use the words as labels; for example, in a table of the SPSS output. When visualizing the word/document occurrence matrix, as we explain below, the words can be adapted for use in Pajek at a later stage.

*4.2.1. Factor analysis*

In order to analyze the word/document occurrence matrix in terms of its latent structure, one may wish to conduct a factor analysis in SPSS. The factor analysis can demonstrate which words belong to which components. Prior to the factor analysis one has to calculate the variance of the variables (the words from the matrix). Words with a variance of zero cannot be used in a factor analysis and therefore need to be left out of the process. (The variance can be computed in SPSS by choosing "Analyze – Descriptive Statistics – Descriptives", then selecting all the words into the right column and then ticking "Variance" under "Options".)

The next step is analyzing the data by means of a factor analysis. Choose "Analyze – Data Reduction – Factor" in SPSS. This step is visualized in Figure 5. Select all the variables in the left column, except the ones with a variance of zero, and select them to the right column. Then, under "Extraction", tick "Scree plot" and undo "Unrotated factor solution". Then, under "Rotation", tick "Varimax" and "Loading plot(s)" and finally, under "Options", tick "Sorted by size" and "Suppress absolute values lower than", which is by default set at larger than .10.

Under Extraction it is additionally possible to manually choose the number of factors. When the output of the factor analysis produces too many factors, it may be advised to manually set the number of factors on, for example, six. More than six factors may be difficult to visualize and interpret through Pajek at a later stage.

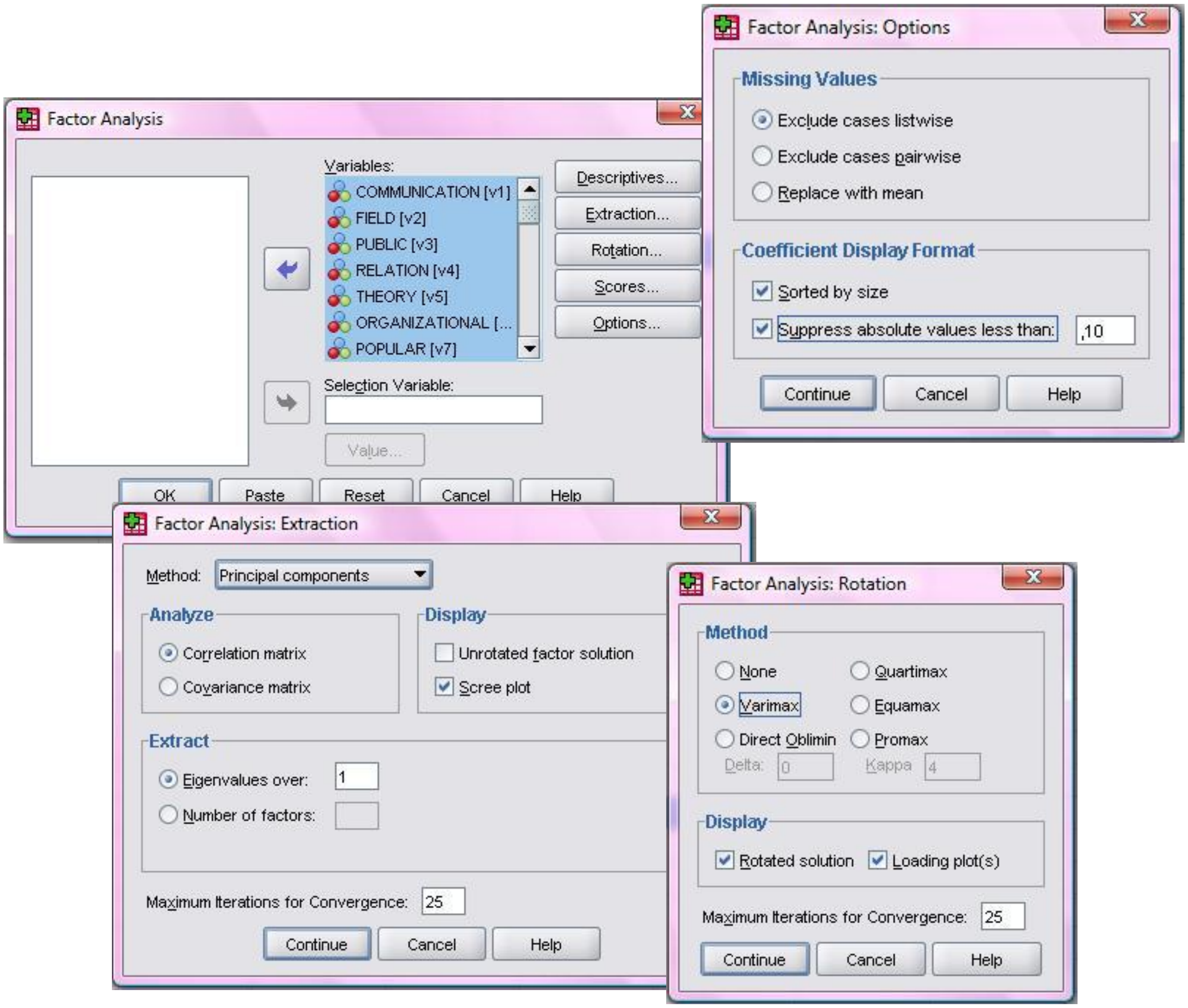

***Figure 5*** Factor Analysis in SPSS

The options are now set in order to conduct a factor analysis. SPSS produces several tables and figures in the output. The most relevant for our purpose is the Rotated Component Matrix. This matrix shows the number of components (factors) and the loading of the different words on the components. At this stage, one can arrange the words under the different components, which can be used when visualizing the word/document occurrence matrix in the next stage. In Figure 6 an example of a few words are visualized with the arrangement under de different components. The various components can be considered as representations of different frames used in these texts. In the example in Figure 6, the texts are built around three different frames. How to use the output to visualize the word/document occurrence matrix will be discussed in the next section.

**Rotated Component Matrix**[a]

| | Component | | |
|---|---|---|---|
| | 1 | 2 | 3 |
| RESEARCH | **,875** | ,436 | -,209 |
| FIELD | **,780** | ,256 | ,572 |
| WITHIN | **,568** | -,674 | -,472 |
| WELL | **,568** | -,674 | -,472 |
| LEVEL | **,332** | | -,940 |
| COMMUNICATION | -,147 | **,968** | ,202 |
| APPLIED | ,533 | **,843** | |
| AREA | ,483 | **,841** | ,243 |
| PUBLIC | ,345 | **,766** | ,542 |
| THEORETIC | ,345 | **,766** | ,542 |
| RELATION | ,345 | **,766** | ,542 |
| TOOLS | ,345 | **,766** | ,542 |
| CONCEPTUAL | ,345 | **,766** | ,542 |
| DEVELOPED | ,585 | **,734** | -,344 |
| CONCLUDES | -,332 | | **,940** |
| YEARS | ,579 | | **,811** |
| ACROSS | ,579 | | **,811** |
| APPLY | ,579 | | **,811** |
| BASED | ,579 | | **,811** |
| TRENDS | ,579 | | **,811** |
| VISUAL | ,324 | -,860 | **,395** |
| STUDIES | -,343 | -,852 | **,395** |

Extraction Method: Principal Component Analysis.
Rotation Method: Varimax with Kaiser Normalization.

a. Rotation converged in 6 iterations.

***Figure 6*** Output Factor Analysis in SPSS (an example with a limited number of words/variables)

In addition to the positive factor loading, one may also wish to take into account that “level” has a negative loading (-.94) on Factor 3.

### *4.2.2. Cronbach's alpha*

Prior to the visualization of the matrix, one may wish to conduct a reliability analysis, by calculating Cronbach's alpha (α) for each frame (component). This measure controls after the factor analysis whether the frames form a reliable scale. First, one can determine which words belong to which frames by using the output of the factor analysis in SPSS, like the example in Figure 6.

The next step is the calculation of Cronbach's alpha in SPSS, by choosing "Analyze – Scale – Reliability Analysis". Select the words from the first frame into the right column and run the reliability analysis by choosing "OK". Figure 7 shows the output of this analysis with Cronbach's alpha for the example from Figure 6, using the second frame which was composed of nine items (that is, words as variables).

**Reliability Statistics**

| Cronbach"s Alpha | N of Items |
|---|---|
| ,949 | 9 |

***Figure 7*** Output reliability analysis (Cronbach's α) in SPSS

In the example in Figure 7, Cronbach's Alpha has a value of .95. In order to guarantee the internal consistency of the scale, Cronbach's Alpha needs to have a minimal value of .65.

### 4.3. Visualizing the word/document occurrence matrix

In this section we explain how to visualize the word/document occurrence matrix by using Pajek[7] and the output of the factor analysis in SPSS. In order to visualize the output of FullText, one is advised to use the file cosine.dat, which was generated by FullText (see chapter 2).[8] In the first part of this section the drawing of the figure is discussed. After that we explain how the visualization can be informed by the output of the factor analysis in SPSS. The final part of this section discusses the layout of the figure and how this can be changed.

Choose "File – Network – Read" to open the file cosine.dat in Pajek. To create a partitioned figure, one can choose "Net – Partitions – Core – All". To draw the Figure, choose "Draw – Draw partition". One can change the layout of the figure by choosing "Layout – Energy – Kamada-Kawai – Free". In this stage, one has created a figure which shows the different components with different colors, as can be seen in Figure 8. However, the algorithm used in Pajek for attributing the colors is different from the results of the factor analysis. We will change this below.

[7] The latest version of Pajek can be downloaded at http://vlado.fmf.uni-lj.si/pub/networks/pajek/ .

[8] The cosine-normalized matrix can be compared to the Pearson correlation matrix which is used for the factor analysis, but without the normalization to the mean. Word-frequency distributions are usually not normally distributed and therefore this normalization to the mean is not considered useful for the visualization. The results of the factor analysis inform us about the latent dimensions which are made visible by the visualization as good as possible. Note that visualization is not an analytical technique.

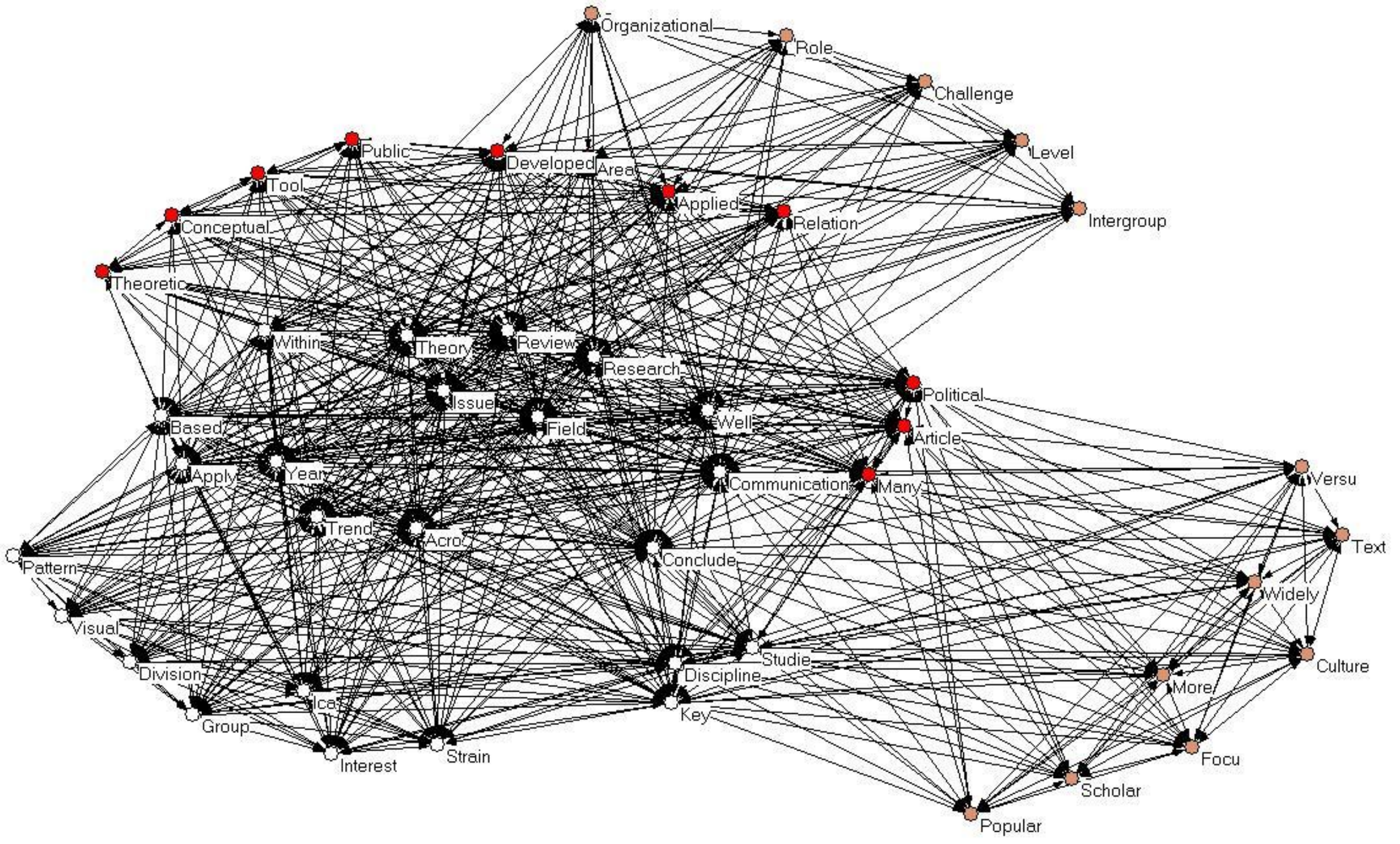

***Figure 8*** Standard Pajek figure with different components

As noted above and shown again in Figure 8, FullText automatically removes an "s" at the end of a word. Also in Pajek it is possible to put back the "s", in case of an incorrect removal. To do so, close the Figure and choose "File – Partition – Edit" in Pajek. In this window one can change the words manually. After closing the window and drawing the partitioned figure again, the words are changed.[9]

The next step in visualizing the word/document occurrence matrix is the adjustment of the figure to the output of the factor analysis in SPSS, as discussed in the previous chapter. After the factor analysis in SPSS, each word was assigned to a specific frame. In the example, there were three different frames made visible in the output (Figure 6). In spite of the fact that Figure 8 also shows three frames in Pajek, there are differences between these frames and the frames from

[9] Alternatively, one can change the words in the input file cosine.dat using an text editor such as Notepad.

SPSS. These differences are being caused by the fact that Pajek uses the cosine matrix while SPSS uses the correlation matrix and performs an orthogonal rotation.

The visualization as shown in Figure 8 can be adjusted to the output of the factor analysis in SPSS. This adjustment can be done in the same way as the changing of the words in the previous section. By choosing "File – Partition – Edit", the frames can be reclassified by assigning the same numbers to words in the same frame.[10] After adjusting the figure from Figure 8 to the factor analysis in SPSS, a new figure can be drawn, which is shown in Figure 9.

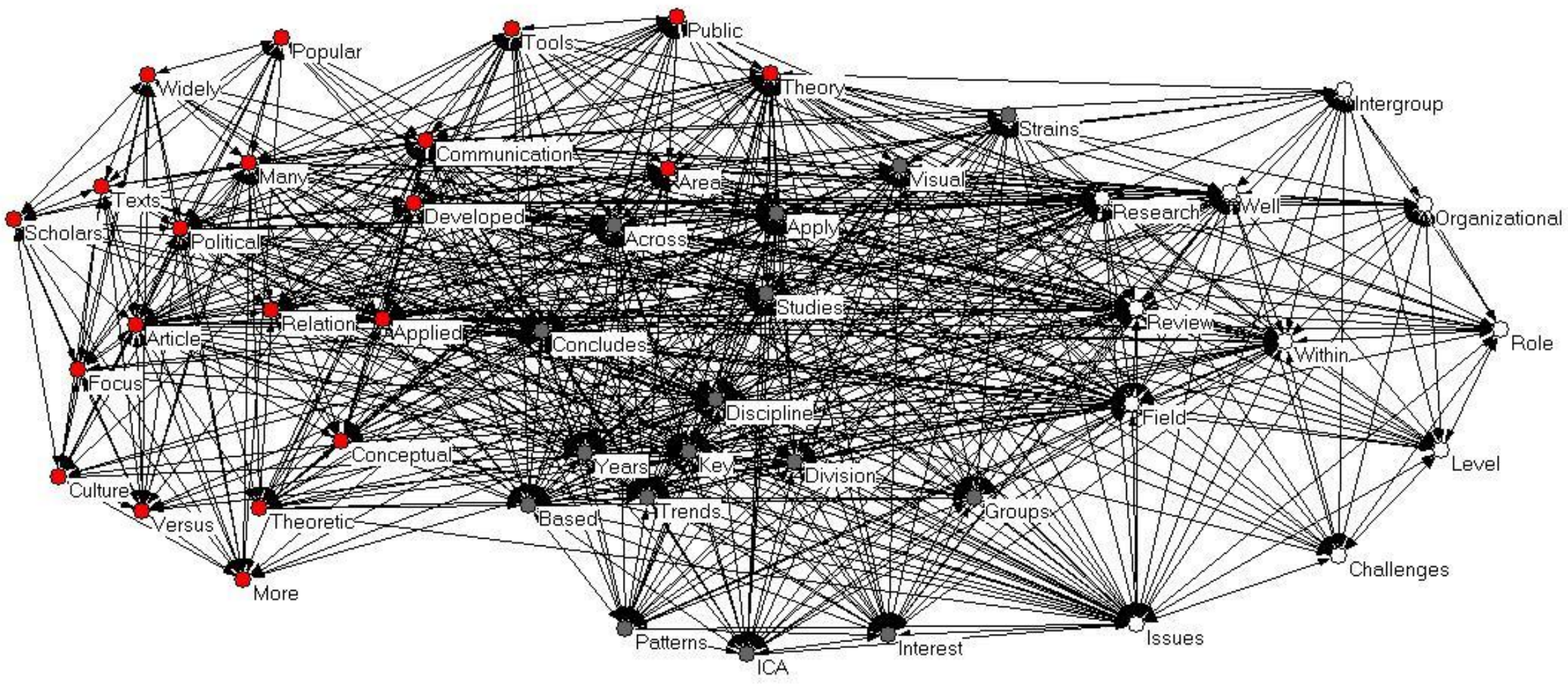

***Figure 9*** Pajek figure adjusted to factor analysis in SPSS

The initial numbers, corresponding to the different frames, are provided by Pajek using another algorithm than factor analysis (the *k*-core algorithm). Nevertheless, the numbers are always one-to-one related to the colours of the vertices in the figure. As can be seen in Figures 8 and 9, it is difficult to read the words in the current layout of the figure. The different lines are also difficult to distinguish.

[10] In the Draw screen, Shift-Left click a vertex to increase its partition cluster number by one, Alt-Left click a vertex to decrease it by one.

There are several options which can increase the readability of the figure. A few of these options are being introduced here. After following these steps, the figure will be better readable and interpretable. Figure 10 provides an overview of the adjustment options in Pajek.

| | |
|---|---|
| ***Background*** | The figure can be read best with a white background. To change the background, choose “Options – Colors – Background” and choose white as the background color. |
| ***Lines*** | To make sure the different lines can be distinguished, it is possible to remove the lines with a value lower than for instance 0.2 (this depends on the figure, different values can be tried). To do so, close the figure, than choose “Net – Transform – Remove –Lines with value – lower than” and fill in the appropriate value. It is also possible to adjust the width of the lines to their values. In order to do so, draw the partitioned figure, then choose the option “Options – Lines – Different Widths”. Since the cosine varies between zero and one, a value of 3 or 5 will provide differences. |
| ***Arrows*** | The arrow heads are not adding anything to the figure, so they can be removed. To do so, choose “Options – Size – of Arrows” and fill in 0. |
| ***Font*** | The size of the font can be changed through “Options – Size – of Font –Select”. Use at least 12 for a PowerPoint presentation in order to make sure the words can be read. To make sure the words do not overlap each other, it is possible to drag the words a little into different directions. |
| ***Vertices*** | The sizes of the vertices can be made proportional to the (logarithm of) the frequency of the words. In order to do so, choose “Options – Size – of Vertices defined in input file”. To enlarge all the vertices, choose “Options – Size – of Vertices” and fill in the size. In Figure 10, the vertices have been given the size of 10. |
| ***Colors*** | To change the colors of the vertices choose “Options – Colors – Partition Colors – for Vertices”. One can change the colors of the vertices, by clicking on the current color and then filling in the number of the wished color as seen on the color pallet. After that, click on OK and close the color pallet. Then click on one of the vertices you want to change and the entire frame will have the wished color. This can be done for each group of vertices. Make sure the colors are in different shades, in order to visually see the differences between the different frames. |

***Figure 10*** Adjustment options in Pajek

Figure 11 shows the same figure as in Figures 8 and 9 after passing through the preceding steps. The words can be read better and the differences in the loadings of the words can be interpreted.

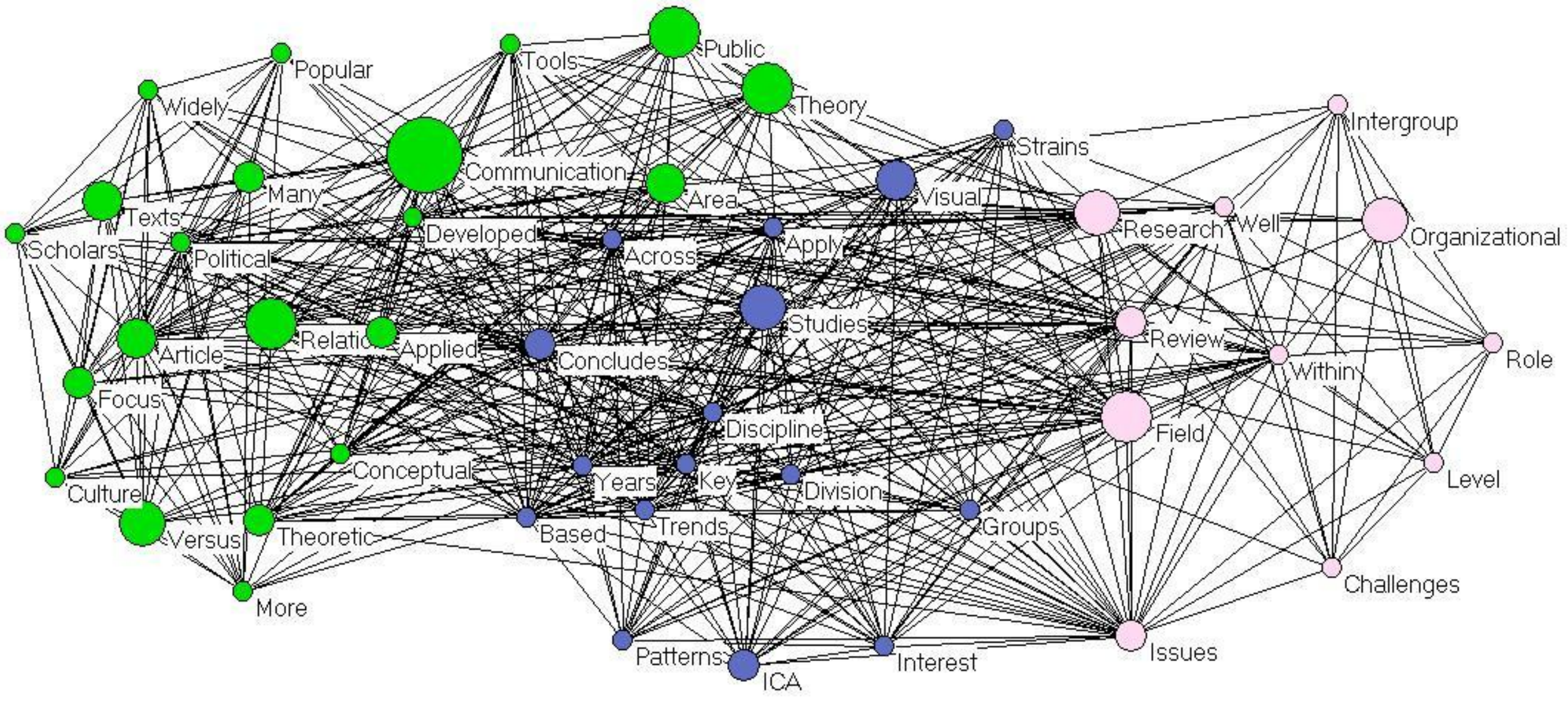

***Figure 11*** Pajek figure after changing the layout

A final option to complete the above figure is the addition of the frames to the figure (using Word or a program like Paint). Through the different words it is possible to name the different frames. Figure 12 shows an example of this addition to the above figure.

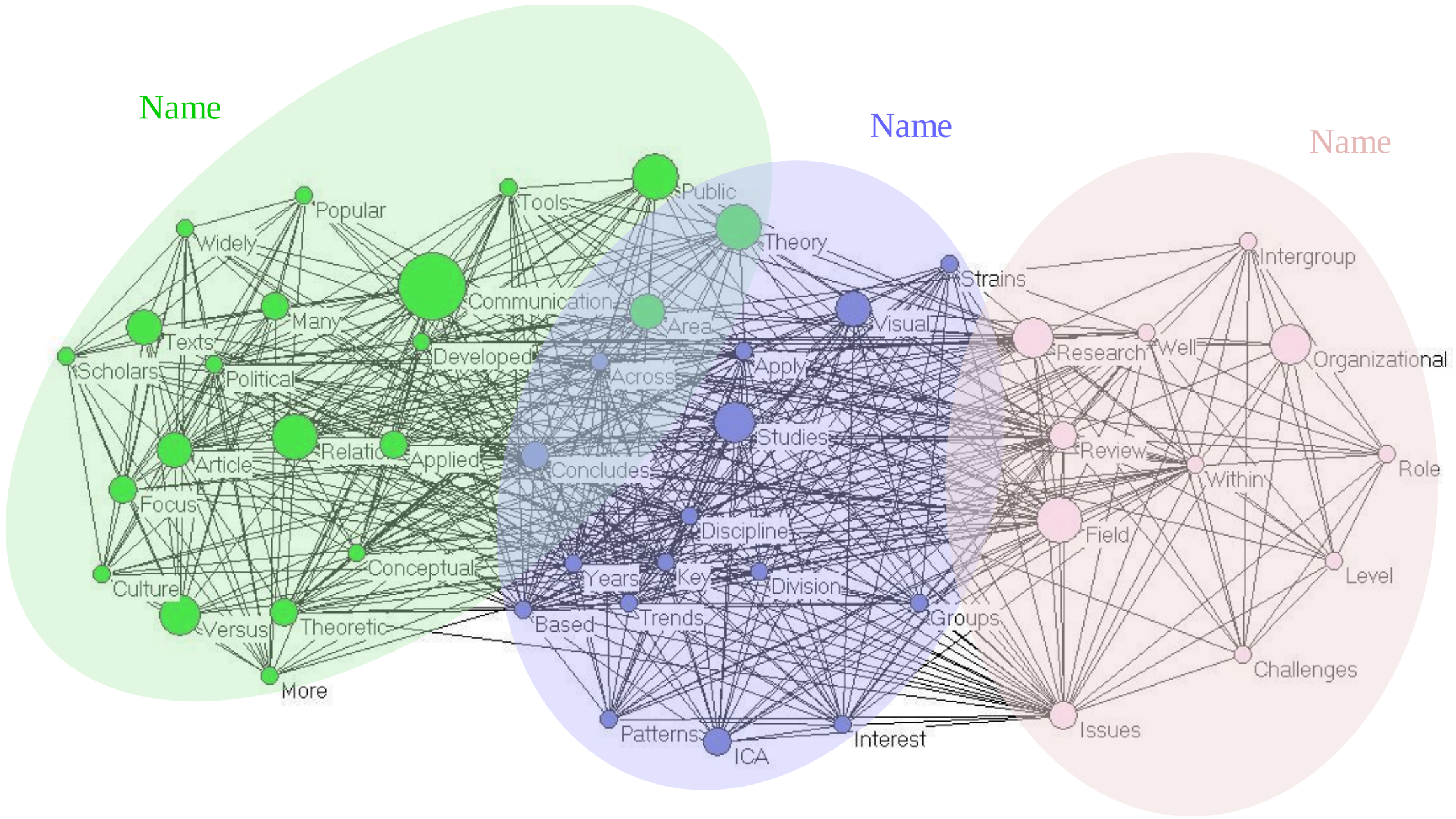

***Figure 12*** Pajek figure after highlighting the different frames

### 4.4. The discourse about "*autopoiesis*" visualized as an example

In this section, a short illustration of the research methodology is provided by visualizing the international discourse on *autopoiesis*. The set of messages consists of nine newspaper articles from various English language newspapers harvested from LexisNexis.[11] The messages were saved as seperate text-files, and the output of the programs Frequency List (FrqList.exe), Full Text (FullText.exe) and SPSS were used to serve as input for Pajek. In Figure 13 the word/document occurrence matrix and the factor analysis are visualized using Pajek.

[11] The Washington Post (2), The Australian (1), Calgary Herald (1), The Herald (Glasgow) (1), The Independent Extra (1) The New York Times (1), The Observer (1), and Prince Rupert Daily News (Britisch Columbia) (1).

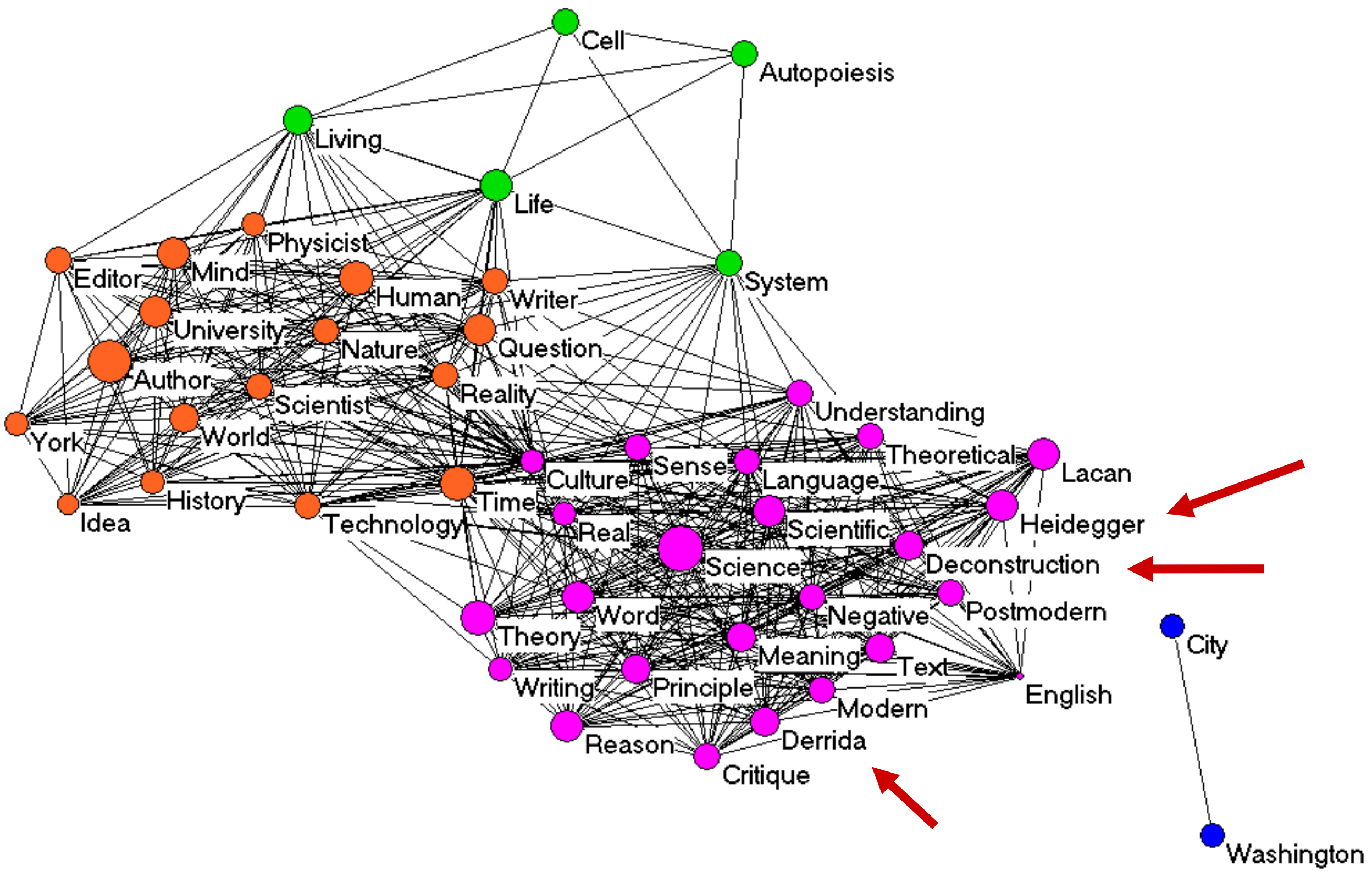

***Figure 13*** Visualization of the discourse on autopoiesis

The visualization of the discourse on autopoiesis provides an illustration of the use of this research methodology. Although it is difficult to draw substantive conclusions based on the above figure – as the sample was only drawn to serve for the purpose of this illustration – one can see for instance that "Derrida" and "Heidegger" are linked semantically via the concept of "Deconstruction." In the same manner it would be possible to study discourses on various concepts, by analyzing the visualizations as provided by Pajek.

## 5. Limitations

Although this method has the advantage of being more objective in frame extraction than other methods, there are of course some limitations as well. First, Matthes and Kohring (2008) argued that all computer-assisted methods in content analysis assume that words and phrases have the same meaning within every context. A human coder could be better able to detect all the different meanings in a text and to provide an interpretation to the contexts of the words. As Simon (2001) also describes, the computer cannot understand human language with all its subtlety and complexiveness. Our methods, however, allow for further extension using, for example, factorial complexity as an indicator of words having different meanings in—translating between—different frames (Leydesdorff & Hellsten, 2005).

Second, one could argue that there is a problem of validity. Some words need to have a large frequency in order to occur in the analysis, in spite of being central to the content of the text (Hertog & McLeod, 2001). This can further be elaborated in terms of various statistics, such as "term frequency-inverse document frequencies" (tf-idf; Salton & McGill, 1983) or the contribution of words (as variables) to the chi-square statistitics of the word/document occurrence matrix. After running ti.exe or fulltext.exe, the file words.dbf contains additional information with these statistics (which are further explained at http://www.leydesdorff.net/software/ti/index.htm; Leydesdorff & Welbers, 2011). Again, this is a problem that occurs as a result of the absence of the researcher in the process of frame abstraction. However, the researcher can manipulate the input file with words (words.txt) which will be used for the analysis.

A final limitation is related to the sources of the media texts. This method can only be applied to texts that are electronically made available (Matthes & Kohring, 2008). As a result of this, visuals or handwritten texts cannot easily be used to study frames using this method. In spite of these limitations, the computer assisted approach of studying frames, as discussed above, provides a relatively researcher-independent assessment of frames, while this objective is hard to accomplish using other methods.

## 6. Future research directions

Our discussion has been oriented towards "getting started" with Pajek for the visualization of latent frames in textual messages. The resulting output can be further embellished for presentation purposes using lesson 6 at http://www.leydesdorff.net/indicators. Other files at this same page use the same techniques for other purposes. For example, one can be interested in the cited references in texts and thus wish to make a citation matrix instead of a matrix of co-occurring words. The basic scheme is that of textual units of analysis (messages) to which a set of variables can be attributed. These variables can be words, author names, institutional addresses, cited references, etc. One can then generate the file matrix.txt and cosine.dat as described above, and use them for analysis in SPSS and/or visualization in Pajek.

In addition to the available statistics in SPSS, Pajek hosts a number of statistics which have been developed over the past few decades in social network analysis. We already mentioned above the *k*-core algorithm which groups together nodes (vertices) which are interrelated with at least *k* neighbours. An introduction to these statistics is provided by: Hanneman, R. A., & Riddle, M. (2005). *Introduction to social network methods*. Riverside, CA: University of California, Riverside; available at http://faculty.ucr.edu/~hanneman/nettext/. An introduction to Pajek is provided by: De Nooy, W., Mrvar, A., & Batagelj, V. (2005). *Exploratory Social Network Analysis with Pajek*. New York: Cambridge University Press.

## 7. Conclusion

In our argument, semantics was considered as a property of language, whereas meaning is often defined in terms of use (Wittgenstein, 1953), that is, at the level of agency. Ever since the exploration of intersubjective "meaning" in different philosophies (e.g., Husserl, 1929; Mead, 1934), the focus in the measurement of meaning has gradually shifted to the intrinsic meaning of textual elements in discourses and texts, that is, to a more objective and supra-individual level (Luhmann, 2002). The pragmatic aspects of meaning can be measured using Osgood *et al.*'s (1957) Likert-scales and by asking respondents. Modeling the *dynamics* of meaning, however, requires further elaboration (cf. Leydesdorff, 2010).

Our long-term purpose is modeling the dynamics of knowledge in scientific discourse. Knowledge can perhaps be considered as a meaning which is more codified than other meanings; it is generated when different meanings can further be compared and thus related (by an observer or in a discourse). As we noted above, meaning can be generated by an observer or in a discourse when different bits of (Shannon-type) information can be related and comparatively be selected. Thus, the selective operation can be considered as recursive. However, the generation of knowledge presumes the communication of meaning (Leydesdorff, 2011). As we have shown, the analysis of the latter requires the progression from the network space to the vector space. The current contribution is made to support the user by facilitating this important step.